\documentclass[letterpaper, 10 pt, conference]{ieeeconf}  
\usepackage{graphicx}
\usepackage{subfigure}
\usepackage{tabu}
\usepackage{booktabs}
\usepackage{enumerate}

\usepackage{times}
\usepackage{epsfig}
\usepackage{amsmath}
\usepackage{amssymb}

\usepackage{textcomp }
\usepackage{graphicx}
\usepackage{subfigure}
\usepackage{color}
\usepackage{multirow}
\usepackage{cite}

\usepackage{algorithm}
\usepackage{algorithmic}

\newcommand{\tabincell}[2]{\begin{tabular}{@{}#1@{}}#2\end{tabular}}

\IEEEoverridecommandlockouts                              

\title{\LARGE \bf
Simultaneous 3D Object Detection and 6-DOF Pose Estimation
}
\author{Hongsen Liu$^*$  
\thanks{*Corresponding author}
\thanks{H. Liu is with the XXXX.}
}

\begin{document}
\maketitle
\thispagestyle{empty}
\pagestyle{empty}

\begin{abstract}
We propose an efficient single shot method for simultaneous 3D object detection and 6-DOF pose estimation in pure 3D point clouds scenes based on a consensus that \emph{one point only belongs to one object}, i.e., each point has the potential power to predict the 6-DOF pose of its corresponding object. Unlike various point clouds based methods for the similar task that converts the point clouds into regular 3D voxel grids to overcome its irregular structure and do voxel-wise prediction, or to segment the set of the target point clouds in advance and predict on the given segmentation, ours is concise enough to solve the point-wise prediction in 3D point clouds without any structure conversion and stepwise processing. The key component of our method is a multi-task segmentation and prediction network, which can simultaneous predicts: (1) the point-wise semantic segmentation for filtering out background points and reducing search space, (2) the 3D locations of the object 3D bounding box vertices for estimating 6-DOF pose transformation and (3) the confidence for evaluating the accuracy of 3D bounding box prediction.

In addition, we design an efficient 3D dataset generation method based on the Augmented Reality technology (AR), and generate expanded training data for two state-of-the-arts 3D object recognition datasets \cite{PLCHF}\cite{TLINEMOD}. We evaluate our proposed methods on the two datasets and the results show that ours can generalize well to multiple scenarios and delivers comparable or surpass performance with the state-of-the-arts.

%

\end{abstract}

\section{INTRODUCTION}
3D object recognition needs to simultaneous detect the given object in a scene and predict its accurate 6-DOF pose, which is of great significance to many practical applications, e.g., robotic manipulation, object modeling and scene understanding. The current methods used for this task can be divided into (1) \textbf{Feature matching methods} and (2) \textbf{CNN-based prediction methods}. In general, feature matching methods needs to generate appropriate features according to the attributes of the given set of object views to build the model database and then match against the scene features. Features can either be the handcrafted features to represent the object surface (texture or surface variation) \cite{LROPS,TLINEMOD,PLCHF} or, more recently, the learning based features \cite{PAEHF,PCONVAE,PVOXELAE}. The performance of such methods mainly depends on the discriminative power of the feature and the coverage of the 6-DOF pose space in terms of viewpoint and scale, which increases the running times linearly. In addition, a multi-stage hypothesis verification process is adopted inevitably to refine the hypothesis.
\begin{figure}[tbp]
\setlength{\abovecaptionskip}{0.0cm}
\setlength{\belowcaptionskip}{-0.0cm}
\center
\label{fig:subfig:a} 
\centering
    \includegraphics[trim = 5mm 0mm 0mm 5mm, clip, width=0.5\textwidth]{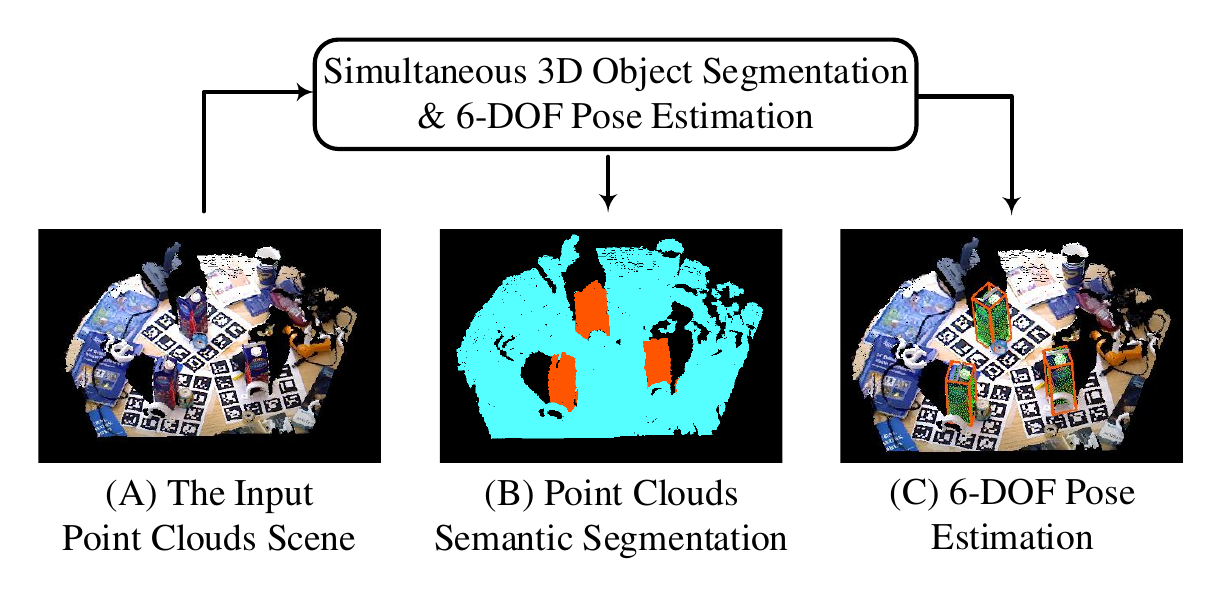}
\caption{ Our method directly operates on the raw point clouds scene and can predict point-wise semantic segmentation and 6-DOF pose estimation. (A) The input point clouds scene, (B) the point-wise semantic segmentaion, (C) the result of the 6-DOF pose estimation with 3D bounding boxes.}
\label{fig:methodview} 
\end{figure}

Recently, a variety of 2D CNN-based methods are proposed to address these limitations \cite{BM,SSD6D,BB8,Seamless}. Instead of using the feature matching engineering, these methods convert the distribution of the whole 6-DOF pose space into a set of network weights based on a large number of training datasets, thus eliminating the linear running time. For example, SSD-6D \cite{SSD6D} depends on the SSD \cite{SSD} architecture to predict 2D bounding boxes and the rough pose information including the possible viewpoint and in-plane rotations, which needs additional hypothesis verification. BB8 \cite{BB8} first segment the interesting objects in 2D and then predict the 2D locations of the projections of the object 3D bounding box for the given segment part, which are then used to compute the 6-DOF pose using a PnP algorithm \cite{PNP}. Seamless \cite{Seamless} relys on the YOLO \cite{YOLO} architecture, which can predict the 2D locations of the projections of the object 3D bounding box without additional processing except using PnP to generate the 6-DOF pose.

\begin{figure*}[htbp]
\setlength{\abovecaptionskip}{-0.cm}
\setlength{\belowcaptionskip}{-0.0cm}
\center
\label{fig:subfig:a} 
\centering
    \includegraphics[trim = 5mm 5mm 0mm 4mm, clip, width=0.82\textwidth]{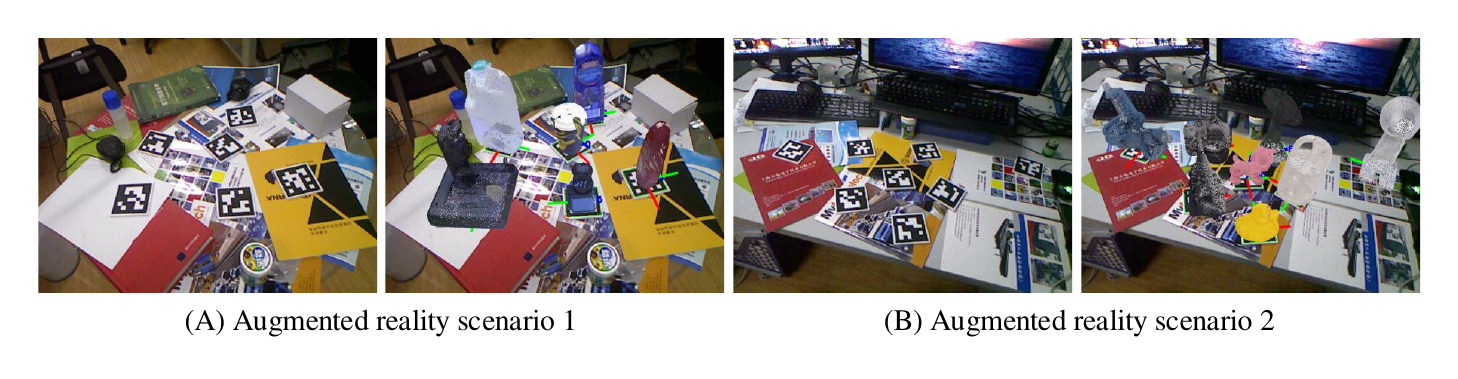}
\caption{The Augmented Reality technology can be used to quickly create 3D object recognition datasets in fixed working scenarios. (A) the generated AR scenario for LC-HF dataset \cite{PLCHF}, (B) the generated AR scenario for LineMod dataset \cite{TLINEMOD}.}
\label{fig:ar} 
\end{figure*}

Most of the above state-of-the-arts cannot directly acquire the 6-DOF pose of the objects in 3D space. In this paper, we propose an efficient single shot method for simultaneously detecting an object in pure 3D point clouds scenes and predicting its 6-DOF pose without complicated post-processing. For 3D point clouds representation, such data is often converted into regular 3D voxel grids, which introduces the voluminous and geometric information loss unnecessarily. Based on a simple consensus that \textbf{one point only belongs to one object}, for a given point clouds scene, each point in the scene has the potential power to predict the 6-DOF pose of its corresponding object. The key component of our method is multi-task segmentation and prediction as shown in Fig.~\ref{fig:methodview}, which can simultaneous predicts: (1) the point-wise semantic segmentation, (2) the 3D locations of the vertices of the object 3D bounding box and (3) the confidence of bounding box prediction. The purpose of the segmentation task is to separate the object points from the background, thus filtering out the unnecessary 3D bounding box prediction. The prediction of the confidence is used to evaluate the accuracy of bounding box prediction and as the score of non-maxima suppression.

In addition, we design an efficient dataset generation method based on Augmented Reality technology (AR) as shown in Fig.~\ref{fig:ar}, and generate expanded training data for two state-of-the-arts datasets LC-HF \cite{PLCHF} and LineMod \cite{TLINEMOD}.  To demonstrate the performance of the proposed method, comparison experiments on these two public datasets are performed. In summary, our main contributions are:
\begin{itemize}
\item We propose a concise enough method for simultaneous 3D object detection and 6-DOF pose estimation, which can do point-wise prediction in 3D point clouds without any structure conversion and stepwise processing.



\item We design an efficient dataset generation method for 3D object recognition based on Augmented Reality technology (AR), which can be used to quickly create 3D object recognition datasets in fixed working scenarios.

\item  We propose extensive experiments to validate the effectiveness of our method using two public datasets, where ours delivers comparable or surpass performance with the state-of-the-arts.
\end{itemize}

\section{RELATED WORKS}
We present a overview of the existing main methods work on 3D object detection and 6-DOF pose estimation. Generally, these methods can be roughly introduced from (1) Feature matching methods to (2) CNN-based prediction methods.

\textbf{Feature matching methods}. Such methods can be divided into handcrafted features and learning features, which the performance is mainly depending on the discriminative power of the feature and the coverage of the 6-DOF pose space in terms of viewpoint and scale. 1) The handcrafted features. 2D local features \cite{ASIFT,SIFT,SURF} are extracted from the 2D RGB image and then back projected to real 3D space. These methods perform well on textured objects but suffer from texture-less. 3D local features \cite{LSPINIMAGE,LSIGNATURE,LSHOT,LTRISI,LROPSCOMPARE} are designed depending on various local 3D surface, which is built by the distribution of local neighboring points. These methods can handle texture-less objects, but are limited to objects with rich variations of the surface normal.
Template features \cite{TLINEMOD,TLINEMODP1,TLINEMODP2} are usually acquired from the object scanning under multiple viewpoint and scale, which is suitable for texture-less objects. LineMod \cite{TLINEMOD} performs robust 3D object detection by embedding quantized image contours and normal orientations on RGB-D images. \cite{TLINEMODP1} optimizes the matching via a cascaded classification scheme and gets 10 times speedup. \cite{TLINEMODP2} proposes a improvement approach based on LineMod templates features via hashing matching.
2) The learning features. A. Doumanoglou et al. \cite{PAEHF} learn patch features via an unsupervised deep Sparse Autoencoder instead of manually designed along with random forests based voting schemes for the estimation of the 6D pose. W. Kehl et al. \cite{PCONVAE} train a Convolution Autoencoder to extract patch features, which gives better performance. Liu et al. \cite{PVOXELAE} present a 3D Autoencoder by converting the point cloud into voxel grids for fully using the 3D spatial structure information.

\textbf{CNN-based prediction methods}.
Recently, CNN-based methods have gradually been used to solve these limitations \cite{SSD6D}\cite{BB8}\cite{Seamless}. SSD-6D \cite{SSD6D} as a pipeline focus on the 2D object detection relies on the SSD architecture \cite{SSD}, which can simultaneous predicts 2D bounding boxes with the object class, the scores for possible viewpoints and the class of in-plane rotations. This is followed by an post-processing to transform 2D bounding boxes to 6-DOF pose hypotheses. BB8 \cite{BB8} is a two-step 3D bounding box detection pipeline, which first segment the object in 2D image and then predict the 2D locations of the projections of the object 3D bounding box for the given segmentation. The 6-DOF pose is computed by using a PnP algorithm \cite{PNP}. Seamless \cite{Seamless} is a different pipeline that relies on the YOLO \cite{YOLO} architecture to predict the 2D projections of the vertices of the object 3D bounding box directly without any post-processing, which are then transform to 6-DOF pose using the PnP algorithm. Most of the above methods can not directly acquire the 6-DOF pose in 3D space of the objects.

\begin{figure*}[tbp]
\center
\setlength{\abovecaptionskip}{-0cm}
\setlength{\belowcaptionskip}{-0.0cm}
\includegraphics[trim = 0mm 6mm 0mm 5mm, clip, width=1.0\textwidth]{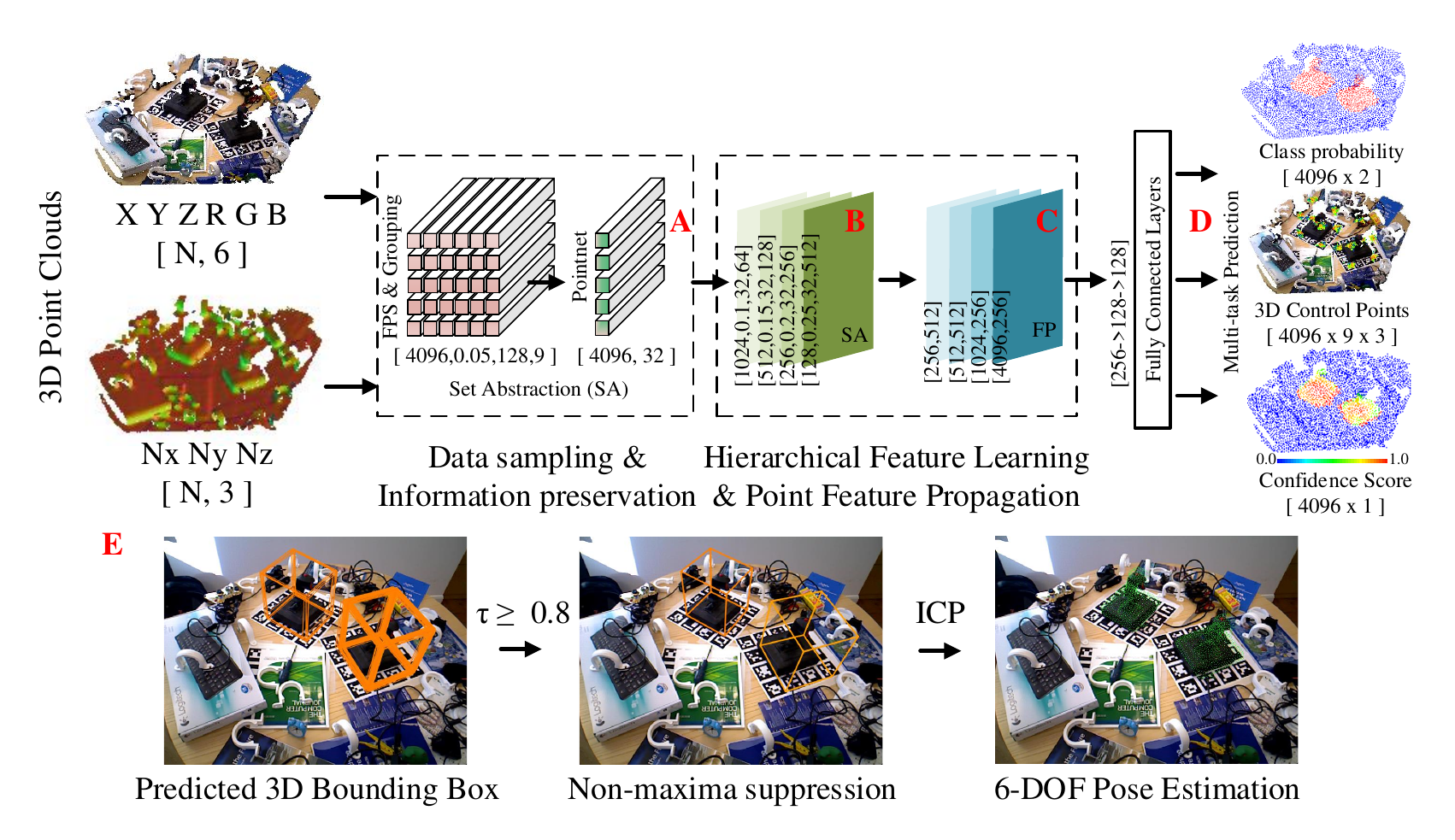}
\caption{The framework of our simultaneous 3D object detection and 6-DOF pose estimation method. For a given 3D point clouds scenes with multiple attributes, we A) sampling/grouping the input scene and generate low density and high dimensional feature data, B\&C) enable to learn point-wise local features via Pointnet \cite{PointNet2} architecture, D) predict the point-wise class probability, 3D control points of object 3D bounding boxes and the confidence score, E) refine the prediction via non-maxima suppression and Iterative Closest Point (ICP) \cite{ICP}.}
\label{fig:framework}
\end{figure*}
\section{OUR MULTI-TASK NETWORK}
The goal of our method is to design an end-to-end trainable network, which can output multi-task predictions for simultaneous 3D object detection and 6-DOF pose estimation without complicated post refine-processing as shown in Fig.~\ref{fig:framework}. The architecture of ours is inspired by Pointnet \cite{PointNet2}, which has a impressive performance of classification, semantic segmentation and part segmentation in point clouds scenes. The architecture of Pointnet is a single-task classification network, which enables to learn point-wise local features by the Set Abstraction (SA) and the Feature Propagation (FP) architecture \cite{PointNet2}. With the powerful local features capturing ability of Pointnet, we use a multi-task loss to predict 1) the class probability for filtering out the background points and judging the instance class, 2) the 3D bounding boxes denoted as 9 control points (8 corners of the box plus 1 center point), 3) the confidence scores for evaluating the accuracy of the bounding boxes prediction. For a given point clouds scene as input, the output of our network is a tensor of dimension $[k\times(9\times3+C+1)]$, where includes $k$ keypoints, $9$ $(x, y, z)$ control points, $C$ class probability and $1$ confidence score. We refine the prediction by Non-maxima suppression and ICP \cite{ICP}, the 6-DOF pose can be computed by solving the spatial transformation between the 3D bounding box of the prediction and the models.

\subsection{Data sampling and information preservation}
Typically low resolution depth sensors have more than 30k points.
In our case, operation at this magnitude points not only beyond the memory capacity on the computing platform, but also the redundant points may bias the detection. Direct sampling the data can definitely lead to information loss, so we need to balance the data sampling and information preservation.

\begin{figure}[htbp]
\setlength{\abovecaptionskip}{-0cm}
\setlength{\belowcaptionskip}{-0.0cm}
\center
\includegraphics[trim = 5mm 5mm 0mm 5mm, clip, width=0.5\textwidth]{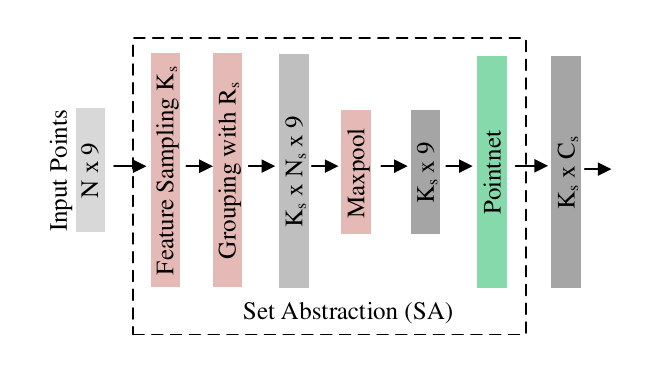}
\caption{The architecture of the Set Abstraction (SA) layer, where contains a sampling layer for sampling keypoints, a grouping layer for finding neighboring points and a Pointnet \cite{PointNet} layer for extracting new features. }
\label{fig:SFEMINI}
\end{figure}

The process of Set Abstraction (SA) of Pointnet \cite{PointNet2} is used to process and abstract a set of input points to produce a new set with fewer elements, which mainly contains: a sampling layer for sampling $K_s$ centroids of local regions, a grouping layer for finding $N_s$ neighboring points around the centroids in a search radius $R_s$ and a Pointnet \cite{PointNet} layer for encoding input features to high abstraction dimension of $C_s$. The architecture details of SA is shown in Fig.~\ref{fig:SFEMINI}. In our case, we use one SA layer for data sampling and generate labels on the fly. For a given training 3D point clouds with a size of $[N\times 9]$, which includes $N$ points and each point contains a normalized feature vector [$X,Y,Z,R,G,B,N_x,N_y,N_z$], including 3D coordinates, colors and normals. We set the $[K_s,R_s,N_s,C_s]$ of SA as $[4096,0.05,128,32]$ to generate a processed training data with a size of $[4096\times 32]$, including $4096$ keypoints and each contains a feature vector with $32$ dimension as shown in Fig.~\ref{fig:framework} $(A)$. This can be regarded as the keypoints selection and feature extraction process, where the SA layer not only decreases the size of the data, but also preserves the local geometric information via local feature abstraction. Since the SA layer uses Farthest Point Sampling (FPS) \cite{FPS} method, which does not change the index of the original data, we can retrieve the labels of each keypoint directly, including the class label with a size of $[4096\times 1]$ and the 3D coordinates of the 9 control points with a size of $[4096\times 9\times 3]$.

\subsection{The multi-task network architecture}
In this subsection, we introduce the training process of our multi-task network. For a given point clouds scene, after the process of data sampling and information preservation as shown in Fig.~\ref{fig:framework} $(A)$, the input of the following network can be denoted as $[b\times k\times c]$, where $b$ represents the batch size, $k$ represents the keypoints size and $c$ represents the input feature channels (in our case, $b = 4$, $k=4096$, $c=32$ ).

The main architecture of the network includes two main phases, the Hierarchical Feature Learning (HFL) and the Point Feature Propagation (PFP) \cite{PointNet2} as shown in Fig.~\ref{fig:framework} (B\&C). In our case, the HFL is composed by four Set Abstraction (SA) layers, which is used to generate global features by progressively abstract larger and larger local regions along the hierarchy. We give the parameter of $[K_s,R_s,N_s,C_s]$ for each SA layer as shown in Fig.~\ref{fig:framework} $(B)$. Here, the parameter $K_s$ of the four SA layers varying from $[1024\to512\to256\to128]$ and the search radius $R_s$ varying from $[0.1\to0.15\to0.2\to0.25]$ accordingly, where this setting takes into account the scale changes of different models, so that it can adapt to different scale models, but not excessive integration of the whole scene global features. We fix the local sampling $N_s$ to $32$ and set the dimension $C_s$ of the output features as $[64\to128\to256\to512]$.

In the HFL phase, the original keypoints is subsampled to generate global features. However, for the task of segmentation, we need point features for all the original keypoints. The Point Feature Propagation (PFP) phase composed of a number of FP layers \cite{PointNet2} is used to propagate the global features from subsampled points to the nearest upsampled points along the hierarchy and generate new local features that incorporate global features. In our case, the PFP is composed of four FP layers with a output size of $[K_f\times C_f]$, where the $K_f$ varying from $[256\to512\to1024\to4096]$ and the $C_f$ varying from $[512\to512\to256\to256]$ as shown in Fig.~\ref{fig:framework} $(C)$.

The final output of the last FP layer is a tensor with a size of $[4096\times 256]$, which means each keypoint has a feature vector of $256$ dimension. Next, we set up three branches and following with Fully Connected Layers to predict 1) the class probability with a size of $[4096\times 2]$, 2) the 3D control points with a size of $[4096\times 9\times 3]$ and 3) the confidence scores with a size of $[4096\times 1]$ respectively as shown in Fig.~\ref{fig:framework} $(D)$.

The composition of the multi-task loss for training the three branches can be denoted as Eq.~\ref{multitaskloss}:
\begin{equation}
\begin{split}
\mathcal{L}&=\lambda_1\mathcal{L}_{cls}(p,p^*)+\lambda_2p^*\mathcal{L}_{offset}(x,x^*)\\
&+\lambda_3p^*\mathcal{L}_{confidence}
\end{split}
\label{multitaskloss}
\end{equation}
Here, the terms of $\mathcal{L}_{cls}(p,p^*)$ represents the classification loss, $p$ and $p^*$ are the predicted class probability and ground-truth label respectively, which is trained by cross entropy. Specially, we do a binary classification due to all of the used datasets include single class objects in each testing scene. The terms of $p^*\mathcal{L}_{offset}(x,x^*)$ represents the  regression loss of the control points, $p^*$ here means the regression loss is activated only for positive class. In the actual, we predict the offset from the keypoints to the 9 control points, which is composed of the 8 corners and 1 center point of the 3D bounding box. We use SmoothL1 loss to train the offset. The term of $p^*\mathcal{L}_{confidence}$ represents the confidence loss for evaluating the accuracy of the control points prediction, where the label of the confidence is generated on the fly. The confidence loss is also valid for positive class. We use the same confidence score function as \cite{Seamless}, where the score function is denoted as Eq.~\ref{scorefunction}:
\begin{equation}
\mathcal{C}_{score}(x)=
\begin{cases}
e^{\alpha(1-\frac{d_{3D}(x)}{d_{th}})},& \text{if $d_{3D}(x)<d_{th}$}\\
0 & \text{others}
\end{cases}
\label{scorefunction}
\end{equation}
Here, $d_{3D}(x)$ represents the mean 3D distance from the predicted offsets to the groundtruth offsets. In our case, the term of $d_{th}$ is set as $0.06m$ to cut-off the monotonically linear function, the sharpness parameter $\alpha=2$. We use the mean-square loss to train the confidence prediction. Specially, we further compute the term of $1-\frac{d_{3D}(x)}{d_{th}}$ as the actually used confidence score, which is in the range of $[0,1]$. We set $\lambda_1$, $\lambda_2$ and $\lambda_3$ simply to $1.0,1.0,5.0$ respectively.
\subsection{6-DOF Pose Estimation}
The final output of the network is a tensor with a size of $[4096\times (9\times 3 +2 +1)]$. For the 6-DOF pose estimation, firstly, we remove the negative class points and the low confidence score points. In our case, we set a confidence threshold $\tau$ as $0.8$. Then, we compute the 3D bounding box corners for each remaining points by the predicted offset values. Next, we group the valid 3D space into equal voxel grids and count the class of the predicted 3D bounding boxes in each grid, where the center of 3D bounding boxes fall into the grids. We remain the 3D bounding boxes of the same class, which has the largest count numbers. Finally, we use non-maxima suppression to reject low confidence score boxes and compute the final 6-DOF pose by solving the spatial transformation between control points of the remaining boxes and the models bounding box. ICP \cite{ICP} is used to refine the matching. We present the estimation process as shown in Fig.~\ref{fig:framework} $(E)$.

\section{EXPERIMENTS}\label{section:Experiments}
In this section, we compare our method with several representative 3D object recognition methods on two public datasets (the LC-HF dataset \cite{PLCHF} and the LineMod dataset \cite{TLINEMOD}), which contains multiple class of objects with different level of occlusion and clutter. We adopt two standard evaluation metrics to compare with different state-of-the-arts. Firstly, we use the F1-Score defined in \cite{PLCHF}, where the estimation is deemed correct if the mean distance $m$ between the true pose $[R,T]$ of model $\mathcal{M}$ vertices and those estimated given the pose $[R_c,T_c]$ is less than $\gamma$ (15\%) of the object diameter \cite{PLCHF}. Secondly, we compare the accuracy of 6-DOF pose using the ADD metric defined in \cite{TLINEMOD}. We take a pose estimate to be correct if the mean distance $m$ is less than $\gamma$ (10\%) of the object diameter \cite{TLINEMOD}. Specifically, for rotationally symmetric objects, the mean distance is computed as Eq.~\ref{add}:

\begin{equation}
m=avg\sum_{x\in \mathcal{M}}\min_{\mathcal{M}}||(Rx+T)-(R_cx+T_c)||\label{add}
\end{equation}


\subsection{Prepare the Training Data}
The training of our network is in pure 3D space and needs both the point-wise semantic label and the 3D bounding box label.
Since the testing datasets do not provide specific training data, and it is impossible to select and process a large-scale 3D natural scene dataset as the training background to cover all the 6-DOF search space, we design a method that can quickly generate specific 3D datasets based on Augmented Reality (AR) technology.
\begin{figure}[htbp]
\setlength{\abovecaptionskip}{-0.3cm}
\setlength{\belowcaptionskip}{-0.0cm}
\center
\label{fig:subfig:a} 
\centering
    \includegraphics[trim = 5mm 0mm 0mm 5mm, clip, width=0.5\textwidth]{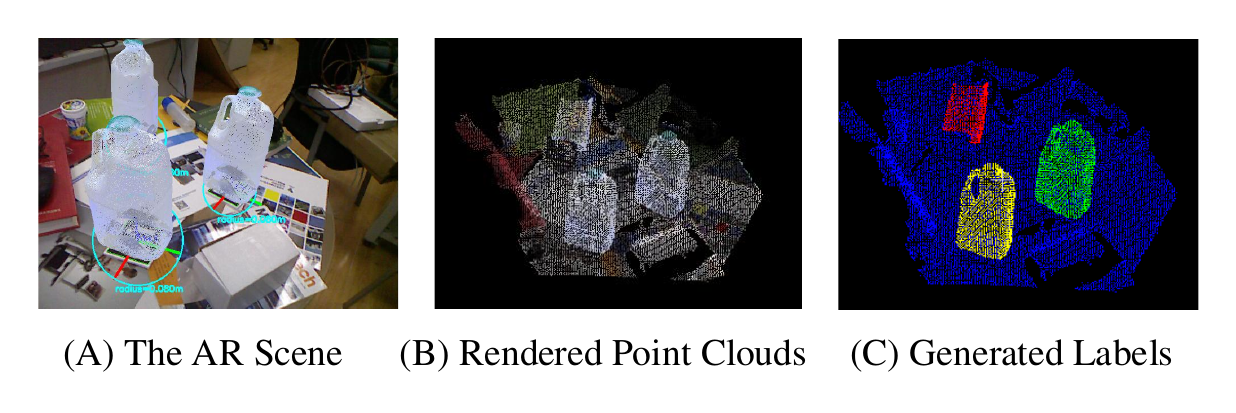}
\caption{ The generation of AR datasets  (A) the AR scene, (B) the rendered 3D point clouds, (C) the generated semantic and instance labels.}
\label{fig:ardata} 
\end{figure}

The method contains three main phases: 1) scenes capture, 2) scenes rendering and 3) labels generation as shown in Fig.~\ref{fig:ardata}.
Firstly, we set up a scene similar to the testing datasets, and then place specific mark boards in it. The method can real-time recognize the transformation between the mark boards and the sensor in the process of collecting the scenes. We use AR technology to visualize the transformed objects mesh models on the mark boards, and display the accurate 6-DOF changes as the sensor moves. In the process of scene capture, we randomly place the mark boards and judge whether the virtual objects are overlapping through the augmented reality scene. For each class of objects, we capture 2k+ frames under a random view and manually remove the images with incorrect pose estimation. Secondly, based on the principle of light propagation and the parameters of real sensors, we render the virtual objects in augmented reality scene, and the occlusion between objects is also presented. Finally, we generate the semantic, instance label and 3D bounding box of each point in the scene, due to the accurate 6-DOF pose of each object is known.

In addition to the generated augmented reality scenes, we sample 10\% of the testing datasets, remove the targets from the scene and replace them with the rendered point clouds of the object mesh models, which are rotated randomly around the axis perpendicular to the working plane every thirty degrees. For each class of object, we first pre-train the weights of our network on the segmentation task same as \cite{PointNet2}, and then train our multi-task architecture on the generated augmented reality datasets, final, use the processed sampled testing datasets to fine-tune the weights.

\subsection{Results On the LC-HF Dataset \cite{PLCHF}}
The LC-HF dataset \cite{PLCHF} contains of 6 objects, where each testing image requires to be detected multiple same targets with different level of occlusion. We use F1-Score as the metric on this dataset and give statistic recognition results as shown in Tab.~\ref{fig:lchfdata}, where the overall average F1-score of our method is the best of 89.2\% in comparison with the other methods. For the camera with lower scores, only a few points are visible in the point clouds space, which leads to the weakly estimation of the 6-DOF pose estimation.
Fig.~\ref{fig:lchfresults} (top row) demonstrates our recognition results on the LC-HF dataset, where shows the prediction of the 3D bounding boxes and the estimated 6-DOF pose. We use the metric Eq.~\ref{add} when evaluating the recognition result
for the rotationally invariant objects \emph{coffee, shampoo, camera and juice} \footnote{We could not find a clear definition of whether a object is rotationally symmetric from other papers, so we explain it here.}.
\begin{table}[htbp]
\renewcommand\arraystretch{0.6}
\setlength{\abovecaptionskip}{-0.0cm}
\setlength{\belowcaptionskip}{-0.5cm}
\centering
\caption{comparison of our method on the re-annotated version of \cite{PCONVAE} with linemod\cite{TLINEMOD}, lc-hf\cite{PLCHF}, convae \cite{PCONVAE} and ssd-6d \cite{SSD6D} in terms of f1-score.}
\setlength{\tabcolsep}{0.8mm}{
\begin{tabular}{l| c c c c c}
\hline
\hline
\specialrule{0em}{2pt}{1pt}
Objects & LineMod\cite{TLINEMOD} & LC-HF\cite{PLCHF} & ConvAE\cite{PCONVAE} & SSD-6D\cite{SSD6D} & OURS\\
\specialrule{0em}{2pt}{1pt}
\hline
\specialrule{0em}{2pt}{1pt}
Coffee   & 0.942                   & 0.891 & 0.972 &\textcolor{blue}{0.983} & \textcolor{red}{0.988}\\

Shampoo  & \textcolor{blue}{0.922} & 0.792 & 0.910 &0.892                   & \textcolor{red}{0.963}\\

Joystick & 0.846                   & 0.549 & 0.892 &\textcolor{red}{0.997}  & \textcolor{blue}{0.958}\\

Camera   & 0.589                   & 0.394 & 0.383 &\textcolor{red}{0.741} &  \textcolor{blue}{0.612}\\

Juice    & 0.595                   & \textcolor{blue}{0.883} & 0.866 &\textcolor{red}{0.919} & 0.871\\

Milk     & 0.558                   & 0.397 & 0.463 &\textcolor{blue}{0.780} &\textcolor{red}{0.961}\\
\specialrule{0em}{1pt}{2pt}
\hline
\specialrule{0em}{2pt}{1pt}
Average  &0.740                    & 0.651 & 0.747 &\textcolor{blue}{0.885} & \textcolor{red}{0.892}\\
\specialrule{0em}{2pt}{1pt}
\hline
\hline
\end{tabular}}
\label{fig:lchfdata}
\end{table}

\begin{figure*}[htbp]
\setlength{\abovecaptionskip}{-0.2cm}
\setlength{\belowcaptionskip}{-0.cm}
\center
\includegraphics[trim = 0mm 3mm 0mm 3mm, clip, width=1.0\textwidth]{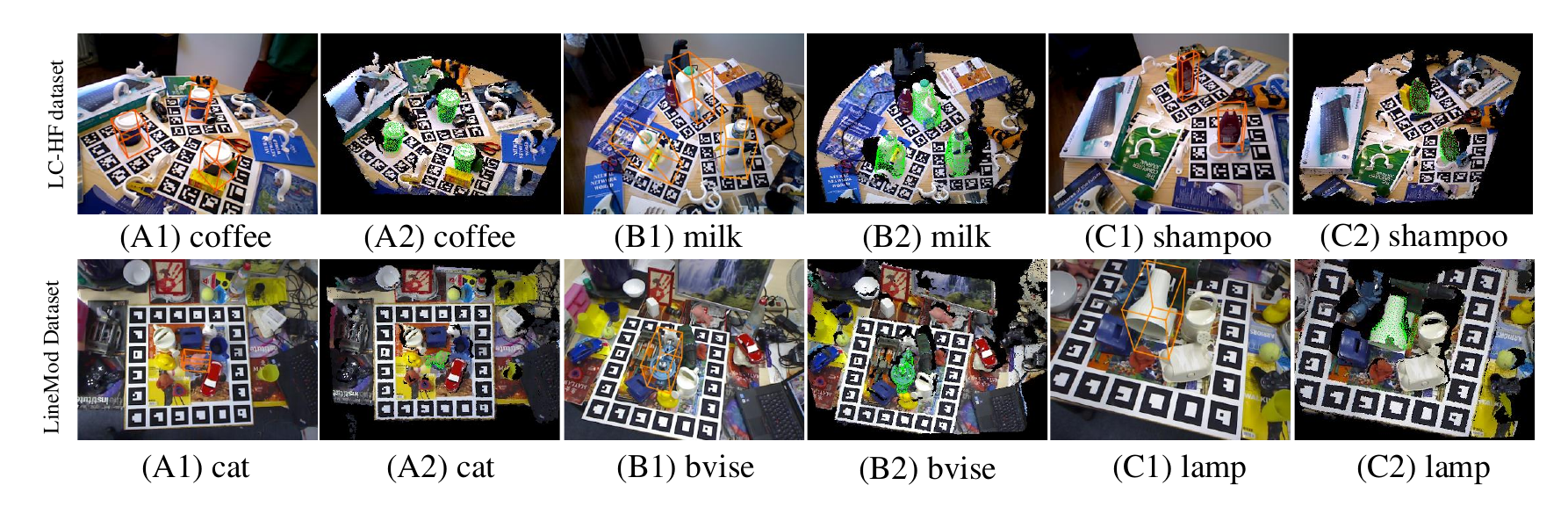}
\caption{Some demo results on the LC-HF dataset \cite{PLCHF} (top row) and the LineMod dataset \cite{TLINEMOD} (bottom row), where the recognition results are shown as the 3D bounding box (A1-C1) and the green transformed model overlaid the estimated location (A2-C2). }
\label{fig:lchfresults} 
\end{figure*}

\begin{table}[htbp]
\renewcommand\arraystretch{0.6}
\setlength{\abovecaptionskip}{-0cm}
\setlength{\belowcaptionskip}{0.4cm}
\centering
\caption{comparison of our method on the linemod dataset \cite{TLINEMOD} with brachmann \cite{BM}, bb8 \cite{BB8}, ssd-6d \cite{SSD6D}, seamless \cite{Seamless} in terms of add metric with/without refinement.}
\setlength{\tabcolsep}{0.46mm}{
\begin{tabular}{l| c c c c | c c c c }
\hline
\hline
\specialrule{0em}{2pt}{1pt}
Methods& \multicolumn{4}{c|}{without refinement} & \multicolumn{4}{c}{with refinement}\\
\specialrule{0em}{2pt}{1pt}
\hline
\specialrule{0em}{2pt}{1pt}
Objects &\tabincell{c}{BB8\\\cite{BB8}} &\tabincell{c}{SSD-6D\\\cite{SSD6D}} & \tabincell{c}{Seamless\\\cite{Seamless}} & \tabincell{c}{OURS\\\ }     &\tabincell{c}{Branchm\\\cite{BM}}      &\tabincell{c}{BB8\\\cite{BB8}}      &\tabincell{c}{SSD-6D\\\cite{SSD6D}} & \tabincell{c}{OURS\\\ }\\
\specialrule{0em}{2pt}{1pt}
\hline
\specialrule{0em}{2pt}{1pt}
ape     &27.9  &0.0    &21.62   &\textcolor{red}{53.2}      &33.2      &40.4    &\textcolor{red}{65.0} &\textcolor{blue}{57.7}  \\

bvise   &62.0  &0.18   &\textcolor{red}{81.80}   &\textcolor{blue}{76.2}      &64.8      &\textcolor{red}{91.8}    &80.0 &\textcolor{blue}{83.3}  \\

cam     &\textcolor{blue}{40.1}  &0.41   &36.57   &\textcolor{red}{55.8}      &38.4      &55.7    &\textcolor{red}{78.0} &\textcolor{blue}{61.5}  \\

can     &48.1  &1.35   &\textcolor{blue}{68.80}   &\textcolor{red}{78.4}      &62.9      &64.1    &\textcolor{blue}{86.0} &\textcolor{red}{96.8}  \\

cat     &\textcolor{blue}{45.2}  &0.51   &41.82   &\textcolor{red}{63.4}      &42.7      &62.6    &\textcolor{red}{70.0} &\textcolor{blue}{66.9}  \\

driller &58.6  &2.58   &\textcolor{blue}{63.51}   &\textcolor{red}{68.8}      &61.9      &\textcolor{red}{74.4}    &\textcolor{blue}{73.0} &71.7  \\

duck    &\textcolor{blue}{32.8}  &0      &27.23   &\textcolor{red}{39.7}      &30.2      &\textcolor{blue}{44.3}    &\textcolor{red}{66.0} &43.7  \\

eggb    &40.0  &8.9    &\textcolor{red}{69.58}   &\textcolor{blue}{62.1}      &49.9      &57.8    &\textcolor{red}{100}  &\textcolor{blue}{70.2}  \\

glue    &27.0  &0      &\textcolor{red}{80.02}   &\textcolor{blue}{74.6}      &31.2      &41.2    &\textcolor{red}{100}  &\textcolor{blue}{80.1}  \\

holep   &42.4  &0.30   &\textcolor{blue}{42.63}   &\textcolor{red}{50.1}      &52.8      &\textcolor{blue}{67.2}    &49.0   &\textcolor{red}{75.8}  \\

iron    &\textcolor{blue}{67.0}  &8.86   &\textcolor{red}{74.97}   &60.6      &\textcolor{blue}{80.0}      &\textcolor{red}{84.7}    &78.0   &65.1  \\

lamp    &39.9  &8.20   &\textcolor{red}{71.11}   &\textcolor{blue}{57.9}      &67.0      &\textcolor{red}{76.5}    &\textcolor{blue}{73.0}   &61.6 \\

phone   &35.2  &0.18   &\textcolor{blue}{47.74}   &\textcolor{red}{79.4}      &38.1      &54.0    &\textcolor{blue}{79.0}   &\textcolor{red}{80.8}  \\

\specialrule{0em}{1pt}{2pt}
\hline
\specialrule{0em}{2pt}{1pt}
Average &43.6  &2.42   &\textcolor{blue}{55.95}   &\textcolor{red}{63.1}          &50.2      &62.7    &\textcolor{red}{79.0}   &\textcolor{blue}{70.4} \\
\specialrule{0em}{2pt}{1pt}
\hline
\hline
\end{tabular}}
\label{fig:add1}
\end{table}

\begin{table}[htbp]
\renewcommand\arraystretch{0.6}
\setlength{\abovecaptionskip}{-0cm}
\setlength{\belowcaptionskip}{-0.4cm}
\centering
\caption{ comparison of our method on the linemod dataset \cite{TLINEMOD} with ssd-6d \cite{SSD6D}, seamless \cite{Seamless} using different threshold for add metric without refinement.}
%
\setlength{\tabcolsep}{1.2mm}{
\begin{tabular}{l| c c c| c  c c}
\hline
\hline
\specialrule{0em}{2pt}{1pt}
Threshold& \multicolumn{3}{c|}{10\%} & \multicolumn{3}{c}{30\%}\\
\specialrule{0em}{2pt}{1pt}
\hline
\specialrule{0em}{2pt}{1pt}
Methods &\tabincell{c}{SSD-6D\\\cite{SSD6D}}  &\tabincell{c}{Seamless\\ \cite{Seamless}} & \tabincell{c}{OURS\\\,} &\tabincell{c}{SSD-6D\\\cite{SSD6D}}  &\tabincell{c}{Seamless\\ \cite{Seamless}} & \tabincell{c}{OURS\\\,}\\
\specialrule{0em}{2pt}{1pt}
\hline
\specialrule{0em}{2pt}{1pt}
Average     &2.42  &\textcolor{blue}{55.95}    &\textcolor{red}{63.1}   &31.65      &\textcolor{blue}{88.25}      &\textcolor{red}{93.47}     \\
\specialrule{0em}{2pt}{1pt}
\hline
\hline
\end{tabular}}
\label{fig:add2}
\end{table}
\subsection{Results On the LineMod Datasets \cite{TLINEMOD}}
The LineMod dataset \cite{TLINEMOD} contains of 13 objects and more than 10k testing images, each testing image has only one target, which are placed with heavy clutter.
We use ADD \cite{TLINEMOD} as the metric on this dataset and give statistic recognition results with or without the process of refinement as shown in Tab.~\ref{fig:add1}, where the overall average ADD score without refinement of our method is 63.1\% outperforming the other state-of-the-arts. We set ICP as the only refinement process to fine matching the predicted results and ours get the second best results except SSD-6D \cite{SSD6D}. This is due to a multi-stage 2D/3D refinement strategy is adopted in SSD-6D. We also compare the accuracy without the refinement using the ADD metric in Tab.~\ref{fig:add2} for different thresholds $\gamma$. When the threshold is set as 30\%, the accuracy of ours attains the best of 93.47\% . In fact, when the threshold is set as 15\%, the accuracy of our method exceeded 90\%. Fig.~\ref{fig:lchfresults} (bottom row) demonstrates our recognition results on this dataset, where shows the prediction of the 3D bounding boxes and the estimated 6-DOF pose. We use the metric Eq.~\ref{add} when evaluating the pose accuracy for the rotationally invariant objects \emph{glue and eggbox} as in \cite{TLINEMOD}.

\subsection{Comparison On the Average Running Time}
We present the average time comparison of our method with  Brachmann \cite{BM}, BB8 \cite{BB8}, SSD-6D \cite{SSD6D} and Seamless \cite{Seamless} as shown in Tab.~\ref{fig:time}, where we record the overall speed time and gives the consumption of the refinement process. Our method achieves the recognition speed of 6fps, in which 113ms is used for point clouds sampling. The refinement process consume 40ms for all objects in the used datasets.

\begin{table}[htbp]
\setlength{\abovecaptionskip}{0cm}
\setlength{\belowcaptionskip}{-0.0cm}
\centering
\renewcommand\arraystretch{0.6}
\caption{compare the average runtime of ours with brachmann \cite{BM}, bb8 \cite{BB8}, ssd-6d \cite{SSD6D} and seamless \cite{Seamless}.}
\scalebox{0.9}{\begin{tabular}{l|l|l}
\hline
\hline
\specialrule{0em}{2pt}{1pt}
Methods   & Overall Speed & Refinement runtime \\
\specialrule{0em}{2pt}{1pt}
\hline
\hline
\specialrule{0em}{2pt}{1pt}
Brachmann \cite{BM}       &2 fps     & 100 ms/object\\
BB8 \cite{BB8}            &3 fps      & 21 ms/object\\
SSD-6D \cite{SSD6D}       &10 fps     & 24 ms/object\\
Seamless \cite{Seamless}  &50 fps     & -\\
OURS                      &6 fps      & 40 ms \\
\specialrule{0em}{2pt}{1pt}
\hline
\hline
\end{tabular}}
\label{fig:time}
\end{table}

We implement the prediction of the network on the Tensorflow framework with a NVIDIA TITAN XP (12GB RAM). The refinement process is executed on a stand PC with a general Intel CPU (i5-3470) at 3.20GHz, 16GB RAM.

\section{CONCLUSIONS}
In this paper, we propose a simultaneous 3D object detection and 6-DOF pose estimation architecture purely in 3D point clouds scenes based on a consensus that one point only belongs to one object, i.e., each point has the potential power to predict the 6-DOF pose of its corresponding object. Ours is concise enough to solve the point-wise 3D object detection and prediction in 3D point clouds, where others need to convert the irregular point clouds into regular voxel grids or to segment the set of the target point clouds in advance and predict on the given segmentation. The various evaluation show that ours can generalize well to multiple scenarios and delivers comparable or surpass performance with the state-of-the-arts. Our method has a casting vote on the potential target centers in the refinement process, which potentially can be used to retrieve the class of each voting point in reverse and obtain additional instance segmentation results. In the future work, we will test the performance of object detection or instance segmentation of our method on public 3D large-scale household datasets.

\bibliographystyle{IEEEtran}
\bibliography{MyBibLib}

\end{document}